\documentclass[10pt,twocolumn,letterpaper]{article}

\usepackage{cvpr}  
\usepackage[accsupp]{axessibility}  

\usepackage{xcolor}
\usepackage{graphicx}
\usepackage{amsmath}
\usepackage{amssymb}
\usepackage{booktabs}
\usepackage{enumitem}
\usepackage{algorithm}  
\usepackage{multirow}  
\usepackage[noend]{algpseudocode}

\definecolor{cvprblue}{rgb}{0.21,0.49,0.74}
\usepackage[pagebackref,breaklinks,colorlinks,allcolors=cvprblue]{hyperref}

\def\paperID{6018} 
\def\confName{CVPR}
\def\confYear{2025}

\title{Knowledge-Aligned Counterfactual-Enhancement Diffusion Perception\\
 for Unsupervised Cross-Domain Visual Emotion Recognition}

\author{Wen Yin$^1$, Yong Wang$^1$, Guiduo Duan$^{1,2}$, Dongyang Zhang$^1$, Xin Hu$^1$, Yuan-Fang Li$^3$, Tao He$^{1,2}$ \thanks{Corresponding author.}\\
$^1$The Laboratory of Intelligent Collaborative Computing of UESTC \\
$^2$Ubiquitous Intelligence and Trusted Services Key Laboratory of Sichuan Province \\
$^3$ Faculty of Information Technology, Monash University \\
{\tt\small \{yinwen1999,cla,guiduo.duan,dyzhang,202411900415\}@uestc.edu.cn}\\
{\tt\small yuanfang.li@monash.edu, tao.he01@hotmail.com}
}
\begin{document}
\maketitle
\begin{abstract}

Visual Emotion Recognition (VER) is a critical yet challenging task aimed at inferring  emotional states of individuals based on visual cues. However, existing works focus on single domains, e.g., realistic images or stickers, limiting VER models' cross-domain generalizability. To fill this  gap, we introduce an Unsupervised Cross-Domain Visual Emotion Recognition (UCDVER) task, which aims to generalize visual emotion recognition from the source domain (e.g., realistic images) to the low-resource target domain (e.g., stickers) in an unsupervised manner. Compared to the conventional unsupervised domain adaptation problems, UCDVER presents two key challenges: a significant emotional expression variability and an affective distribution shift. To mitigate these issues, we propose the  Knowledge-aligned Counterfactual-enhancement Diffusion Perception (KCDP) framework. Specifically, KCDP  leverages a VLM to align emotional representations in a shared knowledge space and guides diffusion models for improved visual affective perception. Furthermore, a Counterfactual-Enhanced Language-image Emotional Alignment (CLIEA) method generates high-quality pseudo-labels for the target domain. Extensive experiments demonstrate that our model surpasses SOTA models in both perceptibility and generalization, e.g., gaining $12$\% improvements over  SOTA VER model TGCA-PVT. The project page is at \textcolor{red}{https://yinwen2019.github.io/ucdver/}. 
\end{abstract}    
\begin{figure}[t]
  \centering
   \includegraphics[width=0.42\textwidth]{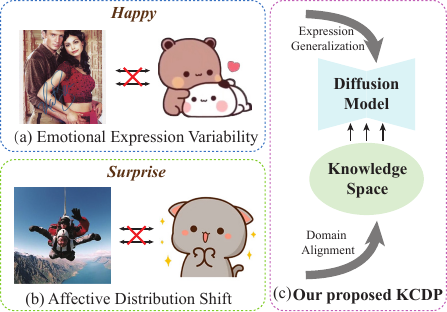}
   \caption{ Illustration of two significant challenges of UCDVER in (a) and (b). Our proposed Knowledge-aligned Counterfactual-
enhancement Diffusion Perception (KCDP)    (in Figure c)  aligns different domains in a shared knowledge space which guides the diffusion model to bridge the domain gap.}
   \label{fig:datashift}
\end{figure} 
\section{Introduction}
\label{sec:intro}


Visual Emotion Recognition (VER), a fundamental task in artificial intelligence and human-computer interaction, aims to identify human emotions through visual cues, such as facial expressions \cite{TextImp2}, body language \cite{BodyLanguage}, and contextual scene features \cite{Scene}. Existing VER methods \cite{ZSER,SAVEA,RER1,RER2,RER3,MDAN,EmoVIT} typically focus on realistic images and have gained considerable advancements on a suite of datasets such as EmoSet \cite{Emoset} and Emotion6 \cite{Emotion6}. 
With the rapid uptake of social media and messaging apps, emojis, stickers, and cartoon characters have been widely used in social media messages. Unfortunately, current VER models cannot handle emotion recognition in these new domains due to the significant emotional expression variability between domains and an affective distribution shift \cite{EmotionalCausality}.

Although pre-trained VLMs such as BLIP \cite{BLIP}, LLaVa \cite{LLaVa}, and InstructBLIP \cite{InstructBLIP}, which have been pre-trained on diverse, multi-domain data, including both realistic images and stickers. However, as demonstrated by our empirical results in Table \ref{tab:UC}, vanilla VLMs underperform in cross-domain visual emotion recognition. We hypothesize that relying solely on VLMs may fail to capture invariant emotional concepts across domains.  Hence, developing an adaptive emotion recognition model that handles universal emotion domains with less adaption cost is important. 

In this paper, we introduce a new challenging task Unsupervised Cross-Domain Visual Emotion Recognition (UCDVER), where a model is trained with labeled source-domain data (e.g., realistic images) and unlabeled target domain data (e.g., stickers), but is employed to recognize emotion in the target domain.
Unlike typical visual adaptive problems such as object detection \cite{DA_OT,DA_OT1} and images classification \cite{DA2,DAMP,UniMoS}, UCDVER needs to solve more severe and sophisticated data-shift problems. Taking the stickers and realistic images as an example shown in Figure \ref{fig:datashift}, two key challenges arise:
\begin{enumerate}[label=\roman*),itemsep=1pt,topsep=0pt,parsep=0pt]
  \item Emotional expression variability: Emotional expressions vary greatly. Realistic images reflect emotions expressed by real humans, while stickers exaggerate or simplify emotions, often focusing on single or multiple virtual elements \cite{Sticker820K, SER30K}.
  \item Affective distribution shift: According to the \emph{Emotional Causality theory} \cite{EmotionalCausality}, an emotion is embedded in a sequence involving (i) external event; (ii) emotional state; and (iii) physiological response. Stickers or emojis emphasize the last two, i.e. (ii) and (iii) \cite{SER30K}, while the emotion in realistic images is often linked to the external context surrounding the subject(s).
\end{enumerate}


The ability to distill a domain-agnostic representation becomes the key to addressing  UCDVER. Inspired by sentiment analysis in NLP \cite{SentimentKnowledge1,SentimentKnowledge2}, we posit that emotions are usually embedded in structured knowledge, e.g., in the form of \textit{subject-verb-object}. For example, in Figure \ref{fig:datashift}, knowledge triplets ``\textit{man - hugging - woman}" and ``\textit{man - is - thrilled.}" can serve as domain-agnostic clues to mitigate emotional distribution shifts. Recently, text-guided conditional Latent Diffusion Model (LDM) \cite{LDM} has been employed to generate cross-domain data \cite{DiffusionGAN3D,LDM2DA,LDM2DG}, exhibiting stunning spatial and semantic consistency and flexibility. Naturally, a question arises: \emph{Can we exploit domain-agnostic cues to guide a pre-trained diffusion model to bridge the severe domain gaps?}

With this question in mind, we propose a Knowledge-aligned Counterfactual-enhancement Diffusion Perception (KCDP) framework, which projects affective cues into a domain-agnostic knowledge space and performs domain-adaptive visual affective perception by a diffusion model.  Specifically, we develop a Knowledge-Alignment Diffusion Affective Perception (KADAP) module which first extracts image captions by BLIP \cite{BLIP} and uses a knowledge parser to extract knowledge triplets as domain-agnostic information. This knowledge guides the LDM in capturing emotional cues by a knowledge-guided cross-attention (KGCA) mechanism. We trained the KGCA by LoRA \cite{LoRA}, which allows efficient fine-tuning by learning low-rank adaptation matrices. To enhance emotion classification, we integrate the features from textual and visual branches. However, our empirical results indicate that a unified global classifier alone does not achieve satisfactory performance. We attribute this to the cross-domain modalities conveying diverse cues for emotion recognition, which a single classifier may fail to fully capture. To this end, we developed a Mixture of Experts (MoE) module to learn invariant information across domains via multiple specialized experts  responsible for capturing different cues in cross-domain modalities. 

To further enhance the performance of the target domain, we introduce a Counterfactual-Enhanced Language-Image Emotional Alignment (CLIEA) strategy to generate higher-quality  pseudo labels for the target domain. In this approach, we first construct a generalized causal graph to represent the causal relationships among elements involved in emotional alignment. To model the indirect effects of various affective prompts on language-image alignment, we employ Counterfactual Contrastive Learning (CCL), which enables the capture of nuances of emotion-specific alignment by isolating the causal impact of prompts. 
In summary, the key contributions of our works are:
\begin{itemize}
    \item We introduce Unsupervised Cross-Domain Visual Emotion Recognition (UCDVER), a new challenging task where a VER  model is trained on a source emotion domain but tested on a new target emotion domain.  
    \item To address UCDVER, we propose a Knowledge-aligned Counterfactual enhancement Diffusion Perception (KCDP)  framework to learn domain-agnostic knowledge across diverse emotion domains.
    \item We develop Knowledge-Alignment Diffusion
Affective Perception (KADAP) and  Counterfactual-Enhanced Language-Image Emotional Alignment (CLIEA) modules align knowledge and visual concepts and generate high-quality affective pseudo-labels.
    \item We establish a benchmark for the task of  UCDVER. Our proposed KCDP framework achieves SOTA performance on various emotion cross-domain scenarios, surpassing several VLMs such as LLaVa.
\end{itemize}

\section{Related Works}
\noindent \textbf{Visual Emotion Recognition.}
The vast majority of previous Visual Emotion Recognition (VER) studies \cite{RER1,RER2,RER3,WEBEmo} have trained and tested the model on a single domain's image, such as EmoSet \cite{Emoset}, WEBEmo \cite{WEBEmo}, Emotion6 \cite{Emotion6}, etc. With the advent of SER60K \cite{SER30K}, Sticker820K \cite{Sticker820K}, a number of arts \cite{SER1,SER3} made good progress in these areas. 
{CycleEmotionGAN++ \cite{CycleGAN}  generated an adapted domain to align the source and target domains at the pixel level with a loss of cycle consistency. In \cite{SfreeVEA}, a new method called BBA was developed that addressed adaptation of the source-free domain for VER.}
However, these efforts do not consider generalization performance in  domains with large distribution shift, e.g., realistic to sticker. 

\noindent \textbf{Diffusion Models for  Visual Tasks.}
Diffusion models are trained as a continuous denoising process from the pure random noise \cite{DDPM}, which is widely used in vision generation \cite{LDM,StableDiffusion,EmoGen,DiffusionGAN3D} and perception \cite{VPD,Difnet,Diffumask} tasks.
Most of these works use text-to-image diffusion models, which guide the reverse process of the diffusion model through prompts. Recent studies \cite{DiffusionCD1,DiffusionCD2,TADP} have explored the use of diffusion models to solve cross-domain visual perception tasks.
\cite{DiffusionCD1} learns the scene prompt on the target domain in an unsupervised way, and uses the prompt randomization strategy to optimize the segmentation head to unlock the domain invariant information to enhance its cross-domain performance. In \cite{TADP}, a diffusion model, as a backbone of the perception model, uses text embeddings after text-image alignment to guide better perception performance. Compared to their use of tedious and irregular texts as guided conditions, we train diffusion models with domain-specific knowledge resulting in a more generalized perception.

\noindent \textbf{Unsupervised Domain Adaptation.}\label{sec:UDA}
UDA has been widely applied across various areas \cite{DAMP,UniMoS,DA_OT,he2019one,song2017deep}, with the goal of transferring knowledge from a labeled source domain to an unlabeled target domain. This approach typically follows two strategies: one is to learn domain-invariant information \cite{DAMP,UniMoS}, and the other is to optimize for domain data discrepancies \cite{DA_SS,DA2}. DAMP \cite{DAMP} leverages domain-invariant semantics through the alignment of visual and textual embeddings in image classification. In \cite{DA_SS}, diffusion models were utilized to generate pseudo-target data to alleviate representation differences between domains. 
While these methods have shown generalization across different domains, variant abstract concepts pose great challenges for conventional UDA  models, particularly the new UCDVER task. Our proposed model  addresses these challenges by effectively capturing domain-invariant information and mitigating data shifts. 

\section{Preliminary}

\noindent \textbf{Problem Statement.}
The Unsupervised Cross-Domain Visual Emotion Recognition (UCDVER) task can be formalized as follows. Given a source emotion domain $(\mathcal{X}_{s},\mathcal{Y}_{s})$ (e.g., realistic images) and an unlabeled target emotion domain $\mathcal{X}_{t}$ (e.g., emojis or stickers images),  the source domain distribution is largely different from the target domain, but their label space is the same. 
The task of UCDVER is to overcome emotion distribution shift so that the model performs well in the target domain.

Suppose that our model is trained with labeled source domain and unlabeled target domain data, and then the trained model is used to make emotion predictions on both domains. Specifically, during the training stage, a model $f$ is optimized based on source domain   $D_s=\{(x_i^s, y_i^s)\}_{i=1}^{n}$ and target domain $D_t=\{x_i^t\}_{i=1}^m$, where $x_i^s \in \mathcal{X}_{s}$ and $x_i^t \in \mathcal{X}_{t}$. At the inference stage, the model predicts the corresponding emotion $\hat{y}=f(x)$ where $x \in D^s \cup D^t$.

\noindent \textbf{Diffusion for Feature Extraction.}
Diffusion models learn the reverse course of the diffusion process to reconstruct the distribution of data \cite{DDPM}. The diffusion process can be modeled as a Markov process:
\begin{equation}\label{eq:diffusion_dist}
  z_t \sim \mathcal{N}\left( \sqrt{\alpha_t} z_{t-1}, (1 - \alpha_t) \mathbf{I} \right),  
\end{equation}
where $z_t$ is the latent variable at time $t$, $\alpha_t$ is the control coefficient. The diffusion model performs the reverse process $p_\theta\left(z_{t-1} \mid z_t\right)$ by predicting noise with an autoencoder $\epsilon_{\theta}$. In the text-to-image LDM,  a language encoder $\tau_{\theta} $ embeds the conditional text $c = \tau_{\theta}(w) $ as the input variable of the denoising autoencoder (typically a U-Net architecture \cite{LDM}). The end of the reverse process is to use a decoder to ensure consistency with the original, such that $\mathcal{D}(\mathcal{E}(x))=\tilde{x}\approx x$. With appropriate reparameterization, the training objective of the LDM can be derived as:
\begin{equation}\label{eq:diffusion_loss}
  \mathcal{L}_{\operatorname{LDM}}=\mathbb{E}_{\mathcal{E}(x),y,\epsilon\thicksim\mathcal{N}(0,1),t}\left[\|\epsilon-\epsilon_\theta(z_t,t,c\|_2^2\right].
\end{equation}

\begin{figure*}[t]
  \centering
    \includegraphics[width=0.97\linewidth]{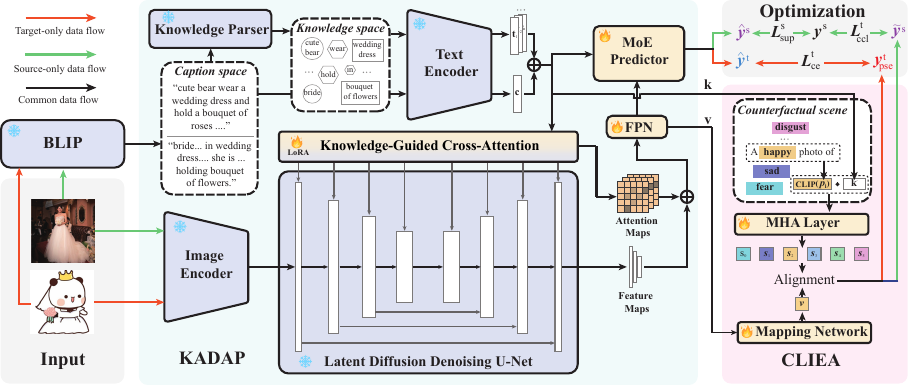}
    \caption{\textbf{Overview of KCDP}, comprising KADAP (\S \ref{sec:KADAP}, \colorbox[RGB]{242,250,251}{green box}), CLIEA (\S \ref{sec:CLIEA}, \colorbox[RGB]{254,236,245}{pink box}) and Optimization (\S \ref{sec:Optimization}, \colorbox[RGB]{242,244,244}{gray box}). Specifically, KADAP uses a BLIP to obtain captions and then employs a knowledge parser to extract knowledge triples. We leverage the CLIP text encoder to encode the knowledge, whose embeddings guide the denoising network via a cross-attention component. The final visual and knowledge representations are fused together to classify emotions via a MoE-based predictor. CLIEA constructs counterfactual samples using knowledge features from causal graphs.  Then a multi-head attention mechanism and a mapping network are used to map linguistic and visual features to the emotional space for alignment to obtain high-quality pseudo-labels of the target domain.
    }
   \label{fig:framework}
\end{figure*}

\section{Methods}
The overall architecture of KCDP is illustrated in Figure \ref{fig:framework}. Briefly, KCDP is composed of two primary modules:  Knowledge-Alignment Diffusion
Affective Perception (\textbf{KADAP}) and Counterfactual-Enhanced Language-Image Emotional Alignment (\textbf{CLIEA}). The \textbf{KADAP} module focuses on learning domain-agnostic knowledge and making robust predictions based on an MoE predictor (\S \ref{sec:KADAP}), while the \textbf{CLIEA} module generates high-quality pseudo-labels for effective training (\S \ref{sec:CLIEA}). In the following sections, we will elaborate on each of them. 

\subsection{KADAP for Domain-agnostic Prediction}\label{sec:KADAP}

KADAP attempts to learn the domain-agnostic cues for UCDVER from the textual and visual perspectives. We elaborate on the following components: textual knowledge extraction, knowledge-guided denoised visual Feature enhancement, and MoE-based emotion prediction.
Specifically, 
given two images $x_s \in \mathcal{X}_{s}$ and $x_t \in \mathcal{X}_{t}$, one of each domain, we first generate the captions $c$ according to the image input $x$ by a VLM (e.g. BLIP) $\mathcal{B}$ and develop a knowledge parser $\mathcal{K}$ to obtain multiple knowledge triples $\{t_i\}_{i=1}^k$. Captions and triplets are encoded by the CLIP text encoder $\mathcal{T}$ as $\mathbf{c}$ and $\{\mathbf{t}_i\}_{i=1}^k$, respectively. An image encoder $\mathcal{V}$ encodes the image into its latent representation $z$. Then, the knowledge features $\mathbf{k}$ obtained by a concatenation of $c$ and $t_i$'s are used as domain-agnostic information to guide the denoising U-Net $\mathcal{D}$ to perceive and enhance visual features $\mathbf{v}$. Finally, an MoE-based Predictor is employed to integrate knowledge features $\mathbf{k}$ and visual features $\mathbf{v}$ to predict emotions $y$.

\noindent \textbf{Textual Knowledge Extraction.}
We first use the large multimodal model InstructBLIP \cite{InstructBLIP} to generate image captions as $c = \mathcal{B}(x)$ by designing effective emotion-related prompts. Subsequently, we exploit the off-the-shell Allennlp Semantic Role Labels (SRL) \cite{AllenNLP} model to identify the semantic roles of each component in the sentence, such as subject and object, etc. We designed an automatic knowledge extraction algorithm to obtain the triples, the details of which are presented in the supplementary material. 
The list of knowledge triples   $\{t_i\}_{i=1}^k \in K$ are fed into CLIP text encoder $\mathcal{T}$ together with the caption $c$ to get their hidden representation as \cite{CKDiffusion}, which can be formalized as $\mathbf{t} = \sum_{i=0}^k \mathcal{T}(t_i)$ and $\mathbf{c} = \mathcal{T}(c)$.
After that, we concatenated them to obtain the final knowledge representation as $\mathbf{k} = \mathrm{concat}(\mathbf{c},\mathbf{t})$  for latent denoising network U-Net.

\noindent \textbf{Knowledge-guided Denoised  Visual  Feature Enhancement.} 
As shown in the centered \colorbox[RGB]{217,229,242}{blue block} of the Figure \ref{fig:framework}, we use a UNet as the denoising network and VQ-VAE \cite{vqvae} as the image initial encoder $\mathcal{V}$, embedding an image into latent $z$. We employ a  cross-attention mechanism to integrate the knowledge into the denoising network as: 
\begin{equation}\label{eq:KDCA}
    \begin{aligned}
    \Phi_i^* = \mathrm{KGCA}(\Phi_i,\mathbf{k}),& i = 1, 2, 3 \cdots,\\
    Q_i = \mathbf{W}_Q^{(i)} \cdot \Phi_i, K_i = \mathbf{W}_K^{(i)}& \cdot \mathbf{k}, V_i = \mathbf{W}_V^{(i)} \cdot \mathbf{k},\\
    \end{aligned}
\end{equation}
where $\Phi_i$ is the $i$-th step hidden representation of the UNet.  

In practice, we froze the parameters of UNet but only finetune  $\mathbf{W}_K^{(i)}$ and $\mathbf{W}_V^{(i)}$. However, our empirical results  show this way does not gain much effectiveness. We conjecture that the main reason lies in the fact that i) certain target emotion  datasets, such as \cite{Art} and \cite{EmotionROI} are data-scarcity and insufficient to train such cross-attention layers and ii) the modification of original parameters would result in the damage of preserved knowledge in the UNet. Inspired by effective finetuning strategy LoRA \cite{LoRA},
we seek to  learn two low-rank matrices instead of fully finetuning the $\mathbf{W}_K^{(i)}$ and $\mathbf{W}_K^{(i)}$. 
Concretely, for the matrix $\mathbf{W}_K^{(i)} \in \mathbb{R}^{d_{in}\times d_{out}}$, we train two decomposition matrices, $\mathbf{B}_i \in \mathbb{R}^{d_{in}\times r}$ and $\mathbf{D}_i \in \mathbb{R}^{r\times d_{out}}$, where $r \ll \mathrm{min}(d_{in},d_{out})$ is the decomposition rank and $i$ represents the level index of the hidden representation in UNet.  The LoRA finetuning process for $\mathbf{W}_K$ can be  calculated as follows:
\begin{equation}\label{eq:LoRA}
    \hat{\mathbf{W}} _K^{(i)}= \mathbf{W}_K^{(i)}+ \Delta\mathbf{W}_K^{(i)} =  \mathbf{W}_K^{(i)}+ \mathbf{B}_i \times \mathbf{D}_i
\end{equation}

Ultimately, we integrate cross-attention maps and intermediate hidden representations at multiple levels of UNet into the enhanced visual feature $\mathbf{v}$ by a Feature Dynamic Network (FPN) \cite{FPN} as:
\begin{equation}\label{eq:VAF}
    A_i = Q_i^{(\Phi)} \times K_i^{(\mathbf{k})}, h_i = F_i  \oplus A_i, \mathbf{v} = \mathrm{FPN}(h_{1,2,3,4}),
\end{equation}
where $F_i$ represents $i$-th level feature in the UNet structure and $\oplus$ represents concatenation operation.

\noindent\textbf{MoE-based  Emotion Prediction.} 
A straightforward approach is to deploy a unified classifier on the visual feature $\mathbf{v}$ and the text embedding $\mathbf{k}$. However, our empirical results suggest that this approach does not yield satisfactory performance, likely due to the diverse information contained within cross-domain modalities for emotion recognition. A single classifier may not effectively capture these varying cues. To address this limitation, we utilize a Mixture of Experts (MoE) architecture to dynamically integrate the visual feature $\mathbf{v}$ and the text embedding $\mathbf{k}$, thereby enhancing the generalization of the fused features and reducing non-essential information within cross-domain modalities. 
Specifically, the MoE module consists of a router $\mathcal{R}(x)$ and $N$ experts $\{\mathcal{E}_i(x)\}^N_{i=1}$, where the router determines which experts to activate. Each selected expert is responsible for capturing a specific type of knowledge within the textual and visual features, enabling a more nuanced and comprehensive representation of emotion recognition.

More specifically, we apply a linear layer to aggregate the {concatenation of the visual feature \(\mathbf{v}\) and the text embedding \(\mathbf{k}\) as \( a = \mathrm{Linear}(\mathbf{v} \oplus \mathbf{k}) \)}. Next, we calculate the routing weights \(\mathbf{W}_E = \{ \mathbf{W}_i \}_{i=1}^N\) for each expert \(\mathcal{E}_i\) through a router \(\mathcal{R}\), as follows:
\begin{equation}
    \mathbf{W}_E = \operatorname{TopK}(\operatorname{Softmax}(\mathcal{R}(a))),
\end{equation}
where \(\mathcal{R}\) projects \( a \) into a 1-D vector, representing the activation probability for each expert. The \(\operatorname{Softmax}\) function then normalizes these weights, highlighting the contribution of the chosen expert. The \(\operatorname{TopK}\) function selects the $k$ most relevant expert by checking the likelihoods of the experts.

Finally, The emotional prediction $y$ is obtained by a weighted sum of the output of $\mathcal{E}$, which is formalized as:
\begin{equation}\label{eq:MoE}
    y = \mathrm{MoE}(a) = \sum_{i=1}^{N} \mathbf{W}_i \mathcal{E}_i(a),
\end{equation}
where $\mathcal{E}_i(a)$ represents the result of the $i$-th expert.

\subsection{CLIEA for Pseudo Labels Generation}\label{sec:CLIEA}
The Counterfactual-Enhanced Language-Image Emotional Alignment (CLIEA) strategy is designed to generate high-quality emotional pseudo-labels for  the target domain. CLIEA is inspired by the causal relationships underlying language-image emotional alignment. We present this strategy through the following steps:
\begin{figure}[t]
  \centering
   \includegraphics[width=0.45\textwidth]{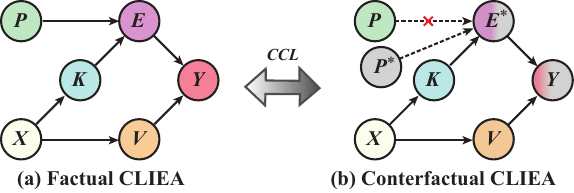}
   \caption{Detail of our CLIEA causal graph. (a) Factual causality $Y_{v,k,p}(X)$. (b) Conterfactual  causality $Y_{v,k,p^*}(X).$}
   \label{fig:casusalgraph}
\end{figure}

Causal graphs are a powerful tool for representing and inferring causal relationships between variables. In our approach, we construct a causal diagram for text-image alignment to analyze the influence of intermediate variables on the final alignment outcome. By intervening on the emotional label prompts, we generate counterfactual samples for use in the latter counterfactual contrastive learning step, allowing the model to achieve a more precise alignment of text and image emotions. We adopt the structured causal model \cite{causalmodel} as our graphical notation. 

 \noindent\textbf{Cause-Effect View Analysis.} 
 As depicted in Figure \ref{fig:casusalgraph}, the causal graph for our proposed CLIEA method illustrates the relationships among five variables: the input image \( X \),  prompt \( P \), the knowledge feature \( K \), the visual feature \( V \), the fused representation \( E \), and the alignment prediction \( Y \). The causal pathways are structured as:
 (1) \( P \to E \to Y \): The affective prompt \( P \) influences the fusion representation \( E \), which in turn impacts the alignment prediction \( Y \).  (2) \( X \to K \to E \to Y \): The image \( X \) is encoded into the knowledge feature \( K \), which then contributes to the emotion embedded in \( E \) after fusion, ultimately influencing \( Y \). (3) \( X \to V \to Y \): The visual feature \( V \) is   derived from the image \( X \) and participates directly in determining \( Y \).

\noindent\textbf{Counterfactual Instantiation.}
Our core idea is to intervene in inferential relationships by constructing counterfactual samples and exploring the influence of different samples on $Y$. 
As shown in the \textit{counterfactual scenario} in Figure \ref{fig:casusalgraph}, we use a diverse emotional prompt \( p^* \), e.g., ``A [Emotion] photo of'', and concatenate it with the knowledge representation $\mathbf{k}$ to generate both counterfactual and factual samples. 
To account for potential biases in the knowledge embedding, we estimate the Total Indirect Effects (TIE)  of the  prompt as follows:
\begin{equation}
\mathrm{TIE} = Y_{v,k,p}(X) - Y_{v,k,p^{*}}(X), 
\end{equation}
where \( Y_{v,k,p}(X) \) and \( Y_{v,k,p^{*}}(X) \) represents the outcome with the original and counterfactual prompt \( p^* \), respectively.  

The next question is how to calculate $Y_{v,k,p}(X)$ and $Y_{v,k,p^{*}}(X)$. A straightforward approach is to leverage the zero-shot capabilities of pre-trained   CLIP \cite{Padclip, DAMP, UniMoS,he2022towards}.  However, in practice, the CLIP model struggles to capture the abstract and complex semantics of emotional expressions, leading to suboptimal alignment between the emotion space and the CLIP space \cite{EmoGen}.
Thus, we leverage a Multi-Head Attention (MHA) mechanism \cite{Attention} to obtain the text prompt embedding by $\mathbf{s}_i = \operatorname{MHA}(\operatorname{CLIP}(p_i) \oplus \mathbf{k})$. A mapping network (an MLP layer) is used to project visual feature $\mathbf{v}' = \operatorname{Linear}(\mathbf{v})$ into the emotional space. Finally, we  calculate the similarity between $\mathbf{v}'$ and $\mathbf{s}_i$ as  $Y_{v,k,p}(X)$:
\begin{equation}\label{eq:Y}
    Y_{v,k,p}(X)=\mathcal{S}(Y_v(X),Y_{k,p}(X)),
\end{equation}
where $\mathcal{S}$ represents the cosine similarity calculation. $Y_{v,k,p^{*}}(X)$ is calculated similarly.

\noindent\textbf{Optimization {by CCL}.}
Intuitively, factual prompt $p$ has a positive impact on the similarity score, while counterfactual prompt $p^*$ negatively affects the text branch due to its incongruity. Hence,  our goal is to  maximize TIE. To this end, we use counterfactual contrastive learning (CCL) technique to push the feature of $p$ away  from the $p^*$ by:
\begin{equation}\label{eq:CCL}
    \mathcal{L}_{ccl} = - \sum\log\frac{\exp(\mathcal{S}(Y_{v_i},Y_{k_i,p_i})/\tau)}{\sum_{j\neq y(i)}^{K}\exp(\mathcal{S}(Y_{v_i},Y_{k_i,p_j})/\tau)},
\end{equation}
where $\tau$ is the temperature coefficient and $K$ is the number of emotional label classes. 

Note that, at inference, we generate  pseudo labels of the target domain by choosing the highest similarity score  with  $K$ affective prompts and knowledge embeddings as:
\begin{equation}\label{eq:pseudolabel}
    y_{pse}^t = \arg\max_{j} \mathcal{S}(Y_v,Y_{k,p_j})
\end{equation}

\subsection{Overall Training Loss}\label{sec:Optimization}
For KADAP, we use a cross-entropy function $\mathrm{CE}(\cdot)$  to calculate the predicted label $\hat{y}$ by the MoE predictor. We use the ground-truth labels $y^s$ for the source domain, while the pseudo-labels $y_{pse}^t$, which is generated by CLIEA, serve as the label for the target domain:
\begin{equation}\label{eq:CE}
    \begin{aligned}
    \mathcal{L}_{ce}^s = \mathrm{CE}(\hat{y}^s,y^s),
    \mathcal{L}_{ce}^t = \mathrm{CE}(\hat{y}^t,y_{pse}^t), 
    \end{aligned}
\end{equation}
Thus, the total loss of our task is:
\begin{equation}\label{eq:loss}
    \mathcal{L} = \lambda_1\mathcal{L}_{ce}^s + \lambda_2\mathcal{L}_{ce}^t + \mathcal{L}_{ccl}
\end{equation}
where $\lambda_1$ and $\lambda_2 $  are weighting factors. Please Refer to \S \ref{sec:implementation} for hyperparameter settings.
\section{Experiments}
\subsection{Datasets}\label{sec:datasets}
We conduct extensive experiments on multiple VER datasets to validate our method.  Specifically, we categorize the datasets into the following domains:

\noindent\textbf{Realistic Images.}
Emoset dataset \cite{Emoset} is a large-scale visual emotion dataset that contains 118,102 images with rich annotation. 
EmotionROI \cite{EmotionROI} collected 1,980 images from Emotion6 \cite{Emotion6} dataset, all from Flickr platform.

\noindent\textbf{Stickers.}
The SER30K dataset \cite{SER30K} collected $30,739$ stickers from $1,887$ themes on the sticker image website. Each sticker is annotated with seven emotional labels. 

\noindent\textbf{Abstract Paintings.} The abstract dataset \cite{Art}, is a collection of $279$ peer-reviewed abstract paintings with no contextual content, composed solely of colors and textures. 

\noindent \textbf{Art Photos.} The artphoto dataset \cite{Art} obtained $806$ art photos using emotion as search terms on the art-sharing website, with the emotional category determined by the artist. 

\noindent\textbf{Universal Images.} The Emo8 dataset \cite{UniEmoX} contains $8,930$ images containing samples of cartoon, nature, realistic, sci-fi, and advertising cover styles.

\subsection{Implementation Details}\label{sec:implementation}
\textbf{Experimental Setup.} 
We establish two experimental settings to evaluate our proposed KCDP:   Domain Adaptation (DA)   and   Universal Cross-domain (UC) setting. The former requires training on the source and target domain, while the latter is solely trained on the source domain and adaptive to other arbitrary domains.  
For the DA setting, we train on the Emoset and SER30K datasets and test using multiple domain datasets. For the UC setting, we only train using the Emoset dataset. 
 
\noindent\textbf{Training Details.} 
Our experiments were conducted using the PyTorch framework \cite{pytorch} on a single $\operatorname{Nvidia Tesla}$ A$800$ GPU. We set the initial learning rate, batch size, decay interval, {$\lambda_1$, and $\lambda_2$} to $1$e-$5$, $16$, $0.01$,  {$1$, and $1$}, respectively, and trained the model for $10$ epochs using the AdamW optimizer. For the image captioner, we used InstructBLIP \cite{InstructBLIP}, which is based on $\operatorname{Flan-t5-xl}$. In the following text, we will use E, R, S, P, A, and U to represent the datasets Emoset, EmotionROI, SER$30$K, abstract, artphoto, and Emo$8$. 

\subsection{Comparison with SOTA Methods}
We compare our model to three types of models: VER models, domain-adaptation models, and VLMs models. The SOTA VER models include MDAN \cite{MDAN}, LoRA-V \cite{SER30K}, and TGCA-PVT \cite{TGCA_PVT}. Specifically, for the DA setting, we compare with the most advanced DA models, namely PADCLIP \cite{Padclip}, AD-CLIP \cite{ADCLIP}, DAMP \cite{DAMP}, and UniMoS \cite{UniMoS}. For the UC setting, we compare with VLM-based SOTA models such as BLIP2 \cite{BLIP}, InstructBLIP \cite{InstructBLIP}, and EmoVIT \cite{EmoVIT}. 
\begin{table*}[htb]
    \centering
    \renewcommand{\arraystretch}{1}
	\setlength{\tabcolsep}{10.85pt}
  \resizebox{0.92\textwidth}{!}{
    \begin{tabular}{l|c|ccc|ccc|c}
        \toprule
         \textbf{Methods}   &\textbf{Type}  & \textbf{E→S}& \textbf{E→P} &\textbf{E→A}&\textbf{S→E} &\textbf{S→P}&\textbf{S→A}&\textbf{Average} \\

        \midrule
        
        PADCLIP (\textit{ICCV'23}) \cite{Padclip}&\multirow{4}*{DA}    & 37.22&25.51&36.84&29.77&26.30&30.45&30.01\\
        AD-CLIP (\textit{ICCV'23}) \cite{ADCLIP}&~                     & 38.07&27.41&35.64&30.52&28.60&31.95&32.03\\
        DAMP (\textit{CVPR'24}) \cite{DAMP}&~                          & 39.61&30.94&36.94&33.76&27.25&31.50&33.33\\
        UniMoS (\textit{CVPR'24}) \cite{UniMoS}&~                      & 38.83&31.27&37.57&34.68&29.36&32.80&34.08\\
        
        \midrule
        
        MDAN* (\textit{CVPR'22}) \cite{MDAN} &\multirow{3}*{VER}   &35.86&30.67&36.25&29.78&26.25&31.45&31.71\\
        LORA-V* (\textit{MM'23}) \cite{SER30K}&~                   &36.21&32.69&33.25&30.87&28.25&35.50&32.79\\
        TGCA-PVT* (\textit{MM'24}) \cite{TGCA_PVT}&~               &37.01&31.90&35.14&30.05&35.14&34.77&34.00\\
        \midrule
        \textbf{KCDP} & DA  & \textbf{62.78}&\textbf{40.55}&\textbf{51.20}&\textbf{41.29}&\textbf{38.69}&\textbf{42.50}&\textbf{46.16}\\
        \bottomrule
    \end{tabular}
    }
    \caption{Experimental results of UCDVER on the state-of-the-art DA and VER models in terms of emotion classification accuracy. * means the model is trained only with the source domain.}
    \label{tab:DA}
\end{table*}
\begin{table*}[htb]
    \centering
    \renewcommand{\arraystretch}{1}
	\setlength{\tabcolsep}{12.6pt}
    \resizebox{0.92\textwidth}{!}{
    \begin{tabular}{l|c|c|ccccc}
        \toprule
        \textbf{Methods}  & \textbf{Backbone}&\textbf{E}& \textbf{R} &\textbf{S}&\textbf{P} &\textbf{A} &\textbf{U} \\
        \midrule
        MDAN (\textit{CVPR'22}) \cite{MDAN}  &   \multirow{3}*{Resnet}    & 75.75&43.34&35.86&30.67&36.25&31.45\\
        LORA-V (\textit{MM'23}) \cite{SER30K}   & ~                    & 76.67&48.18&36.21&32.69&33.25&25.50\\
        TGCA-PVT (\textit{MM'24}) \cite{TGCA_PVT}   & ~                & 78.70&49.74&37.01&31.90&35.14&34.77\\
        \midrule
        BLIP2 (\textit{ICML'23}) \cite{BLIP}&    \multirow{5}*{VLM}   & 49.38&50.51&42.84&29.77&36.25&31.45\\
        LLaVa (\textit{NIPS'23}) \cite{LLaVa}  &  ~                   & 45.07&47.19&39.64&20.52&38.86&33.95\\
        InstructBLIP$\clubsuit$ (\textit{NIPS'23}) \cite{InstructBLIP}  & ~      & 47.28&46.13&38.94&33.76&30.59&28.50\\
        InstructBLIP$\spadesuit$ (\textit{NIPS'23}) \cite{InstructBLIP}  & ~      & 49.51&48.39&40.67&34.50&28.25&30.31\\
        EmoVIT (\textit{CVPR'24}) \cite{EmoVIT}    &                & \textbf{84.02}&53.87&45.66&34.68&45.14&36.89\\
        \midrule
        \textbf{KCDP} &VLM                                       & 83.38                    &\textbf{55.75}&\textbf{57.71}&\textbf{35.84}&\textbf{47.14}&\textbf{47.84}\\
         \bottomrule
    \end{tabular}
    }
    
    \caption{Experimental results of Universal Cross-domain (UC) setting on multiple datasets in terms of emotion classification accuracy. All methods are trained on Emoset (\textbf{E}). $\clubsuit$ and $\spadesuit$ represent the prompt-based and features-based classification strategies for VLMs, respectively.}
    \label{tab:UC}
\end{table*}

\noindent\textbf{Results on the DA setting.}
Table \ref{tab:DA} presents the experimental results for the DA setting, with the arrow ``$\xrightarrow{}$'' indicating the direction from the source domain to the target domain.  
The results show that KCDP outperforms all previous models including all DA and VER models. For the DA models, our method is on average $14$\% better than the SOTA domain adaptation UniMoS when Emoset is the source domain. For SER30K as the source domain, the improvement is on average $9$\% over TGCA-PVT. We deem that the possible reason is the Emoset is more large-scale and enriches with more comprehensiveness. 

\noindent\textbf{Results on the UC setting.}
Table \ref{tab:UC} presents the experimental results of our KADAP in the UC setting, where Emoset is used as the source set but validated on multiple domain datasets. 
Notably, for the VLMs, we devised two classification strategies: prompt-based and feature-based. The former is to design effective emotion-related prompts for VLMs to recognize emotion labels. The feature-based strategy extracts multimodal features by VLMs and trains a unified classifier for all domains.  Our method achieves new SOTA performance and outperforms previous models, e.g., EmoVIT,  across domains by approximately $3$\%. Especially in the widely used Emo8 dataset, we lead previous works TGCA-PVT by $11.39$\%. 
The above empirical analysis indicates that KCDP, by incorporating emotion-related knowledge as domain-agnostic information, significantly enhances the model's domain generalization ability.

\subsection{Ablation Studies}
In this section, we conduct ablation studies on key modules of our KCDP. The modules examined include:   CLIEA,   LoRA fine-tuning strategy, conditional information, and   MoE components.
\begin{table}
    \centering
    \renewcommand{\arraystretch}{1}
	\setlength{\tabcolsep}{14pt}
    \begin{tabular}{l|c|c}
    \midrule
        \textbf{Module} & \textbf{E→S} & \textbf{S→E}\\
        \midrule
         CLIP (vanilla)& 32.07 & 26.53\\
         \midrule
         CLIEA w/ MHA& 57.40 & 39.51 \\
         
         CLIEA w/ MLP& 58.47 & 39.24 \\
         
         CLIEA w/ MHA + MLP & \textbf{62.78} & \textbf{41.29}\\
         \midrule
    \end{tabular}
    \caption{Ablation results of MHA and MLP of CLIEA and comparison with the vanilla CLIP.}
    \label{tab:clieaab}
\end{table}

\noindent\textbf{Effectiveness of CLIEA.}
As shown in Table \ref{tab:clieaab}, we evaluate the effectiveness differences between CLIP and CLIEA on E→S and S→E settings and conduct ablation experiments on MHA and MLP components. Intuitively, using the vanilla CLIP directly for predicting emotional alignment performs poorly. We attribute this to the lack of emotion-related alignment capabilities from pre-trained CLIP. For our CLIEA, introducing either MHA or MLP allows the features to be mapped to a new emotion space instead of the original CLIP space, thus achieving significantly better alignment performance.

\begin{figure}[t]
  \centering
   \includegraphics[width=0.40\textwidth]{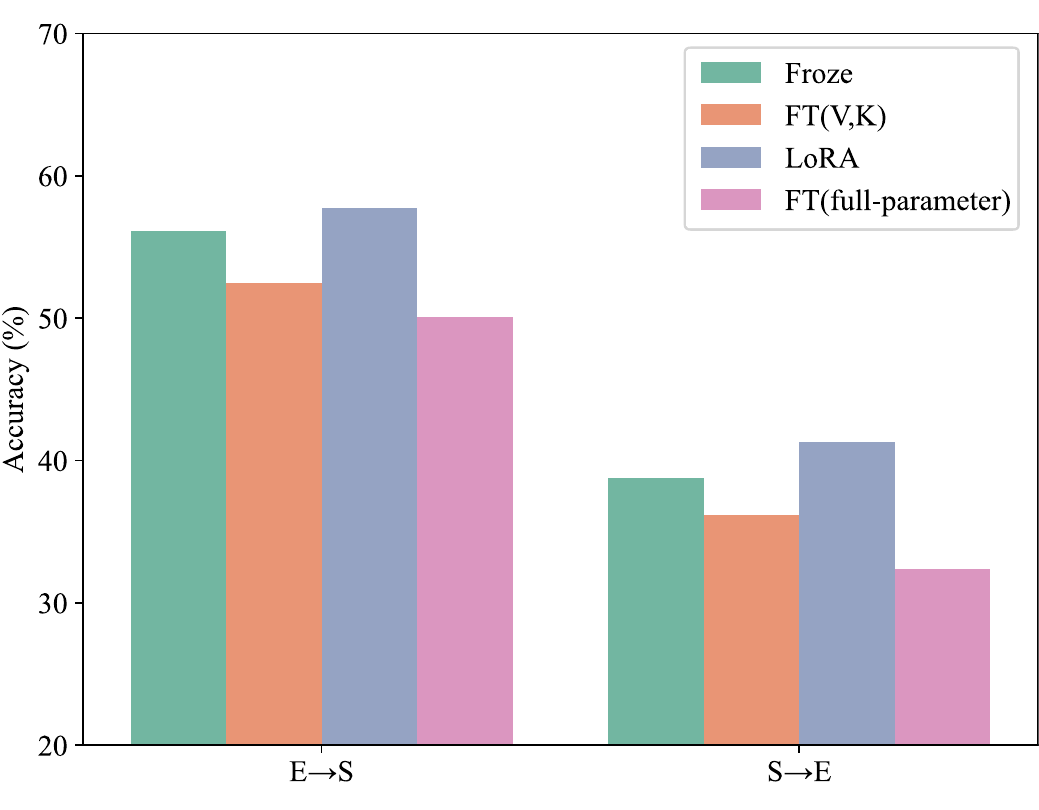}
   \caption{Effectiveness of different fine-tuning strategies on the E→S  and S→E task. Notably, we only report the results on the target domain.}
   \label{fig:loraab}
\end{figure}

\noindent\textbf{Effectiveness of LoRA Finetuning.}
We analyzed four fine-tuning strategies for diffusion models: freezing parameters, fine-tuning $V$ and $K$, LoRA fine-tuning, and full parameter fine-tuning. As shown in Figure \ref{fig:loraab}, we evaluated the accuracy of these four strategies on both the E→S task and S→E task. {We observe that as the extent of parameter adjustment increases, the performance does not gain relative improvements but shows a degradation. We deem that possibly because finetuning many parameters damages the preserved knowledge in the VLMs and degrades their generalization abilities. On the other hand, fully freezing the denoising network makes the extracted feature not compatible with VER tasks. The best way  LoRA can not only keep the pre-trained parameters unaltered but also train two low-rank matrices adept to VER.} 


\noindent\textbf{Effectiveness of Conditions.}
As shown in Table \ref{tab:Ablation}, we conducted an ablation study on the input conditions, specifically the image caption $\mathbf{c}$ and the extracted knowledge triples $\mathbf{t}$, to assess their impact on model performance. Regardless of whether a global classifier or the MoE model was used, the results demonstrate that the model achieves higher accuracy when using $\mathbf{t}$ as a condition compared to $\mathbf{c}$.
Notably, the introduction of $\mathbf{t}$ leads to a significant performance improvement in the target domain, suggesting that $\mathbf{t}$, as domain-agnostic information, is more effective in guiding the denoising network. Moreover, when   $\mathbf{c}$ and $\mathbf{t}$ are combined, the model gains the best results.

\begin{table}[t]
    \centering
    \begin{tabular}{cc|c|cc}
        \toprule
        \multicolumn{3}{c}{\textbf{Modules}}&\multicolumn{2}{|c}{\textbf{Datasets}}\\
        \midrule
        $\mathbf{c}$& $\mathbf{t}$&Classifier&S  &E\\
        \midrule
        \checkmark&-&$\textbf{C}_{global}$ & 72.36  &35.95\\
        -&\checkmark&$\textbf{C}_{global}$ & 72.53(\textcolor{orange}{+0.17})  &36.64(\textcolor{orange}{+0.68})\\
        \checkmark&\checkmark&$\textbf{C}_{global}$ & 72.78(\textcolor{orange}{+0.42})  &36.77(\textcolor{orange}{+0.82})\\
        \midrule
        \checkmark&\checkmark&$\mathrm{MoE}_{\mathbf{v}}$ & \textbf{72.97}(\textcolor{orange}{+0.61}) &\underline{37.49}(\textcolor{orange}{+1.54})\\
        \checkmark&-&$\mathrm{MoE}_{\mathbf{v}+\mathbf{k}}$                     & 72.81(\textcolor{orange}{+0.45})                              &36.66(\textcolor{orange}{+1.16})\\
        -&\checkmark&$\mathrm{MoE}_{\mathbf{v}+\mathbf{k}}$                     & 72.82(\textcolor{orange}{+0.46})                              &37.60(\textcolor{orange}{+1.65})\\
        \checkmark&\checkmark&$\mathrm{MoE}_{\mathbf{v}+\mathbf{k}}$                     & \underline{72.84}(\textcolor{orange}{+0.48})                              &\textbf{37.77}(\textcolor{orange}{+1.82})\\
        \bottomrule
    \end{tabular}
    \caption{Ablation study of Domain Adaptation (DA) setting in S→E task. The ``$\mathbf{c}$'' and ``$\mathbf{t}$'' represent the conditional information captions and triples for the diffusion, respectively. The $\textbf{C}_{global}$ is a global classifier. The $\mathbf{v}$ and $\mathbf{k}$ below MoE represent inputs for visual representation and knowledge representation, respectively.}
    \label{tab:Ablation}
\end{table}

\noindent\textbf{Effectiveness of MoE Components.}
As shown in Table \ref{tab:Ablation}, we initially tested a simple global classifier, denoted as $\textbf{C}_{global}$, as a baseline for this experiment. The results indicate that the performance of $\textbf{C}_{global}$ achieved $72.78$\% and $36.77$\% on the two domains, respectively. However, upon switching to a MoE-based predictor, a significant performance improvement was observed.
Subsequently, we conducted an ablation study on the input components, $\mathbf{v}$ and $\mathbf{k}$, of the MoE module to identify the key inputs influencing emotion prediction. When both the visual representation and the knowledge representation ($\mathbf{v} + \mathbf{k}$) were integrated into the MoE model, the performance on the Emoset dataset was improved further. 


\subsection{Visualized analysis of Embedding}
\begin{figure}[t]
    \centering
    \begin{subfigure}{0.22\textwidth} 
        \centering
        \includegraphics[width=\textwidth]{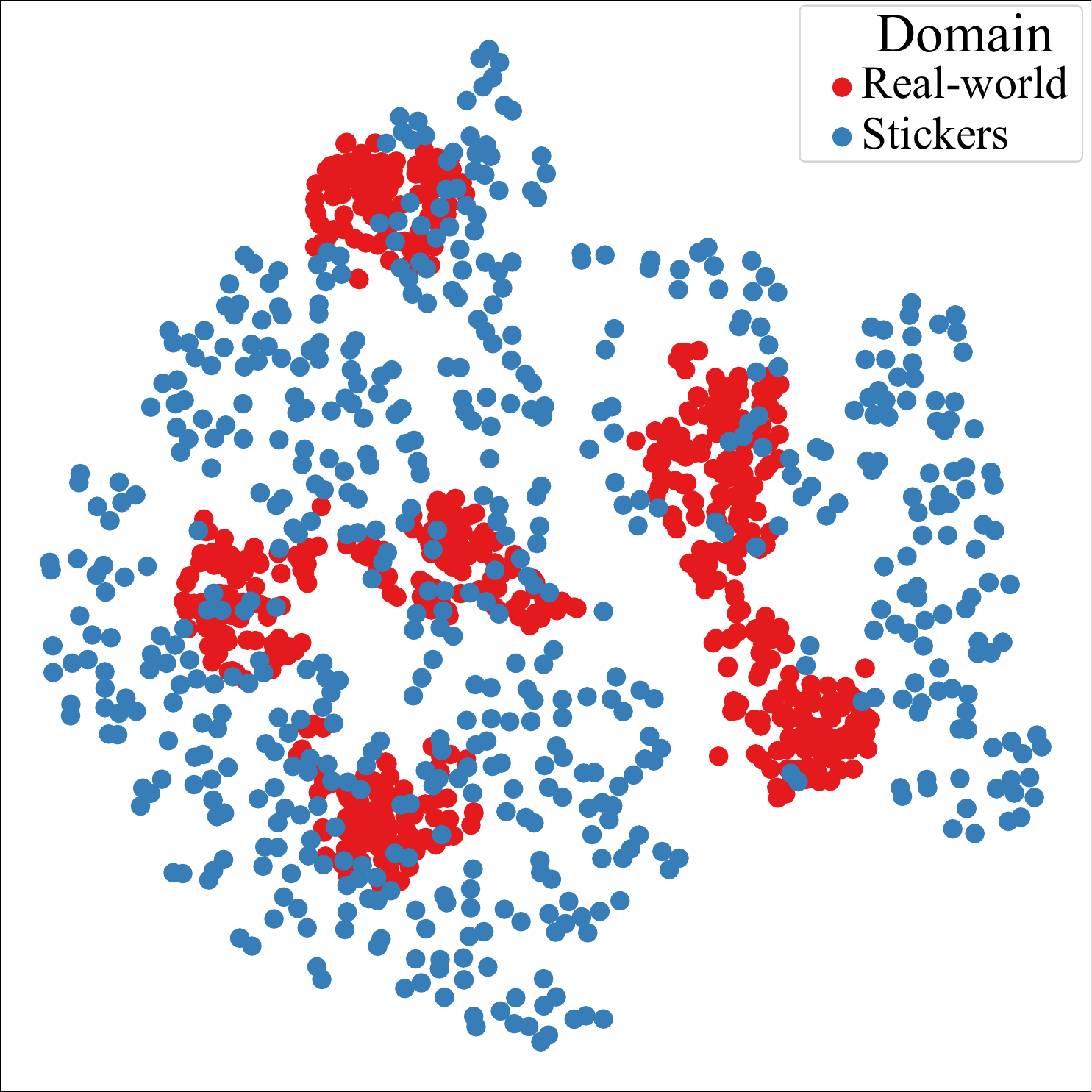} 
        \caption{TGCA-PVT on E→S} 
        \label{subpic:E2S_tpca}
    \end{subfigure}
    \hfill 
    \begin{subfigure}{0.22\textwidth}
        \centering
        \includegraphics[width=\textwidth]{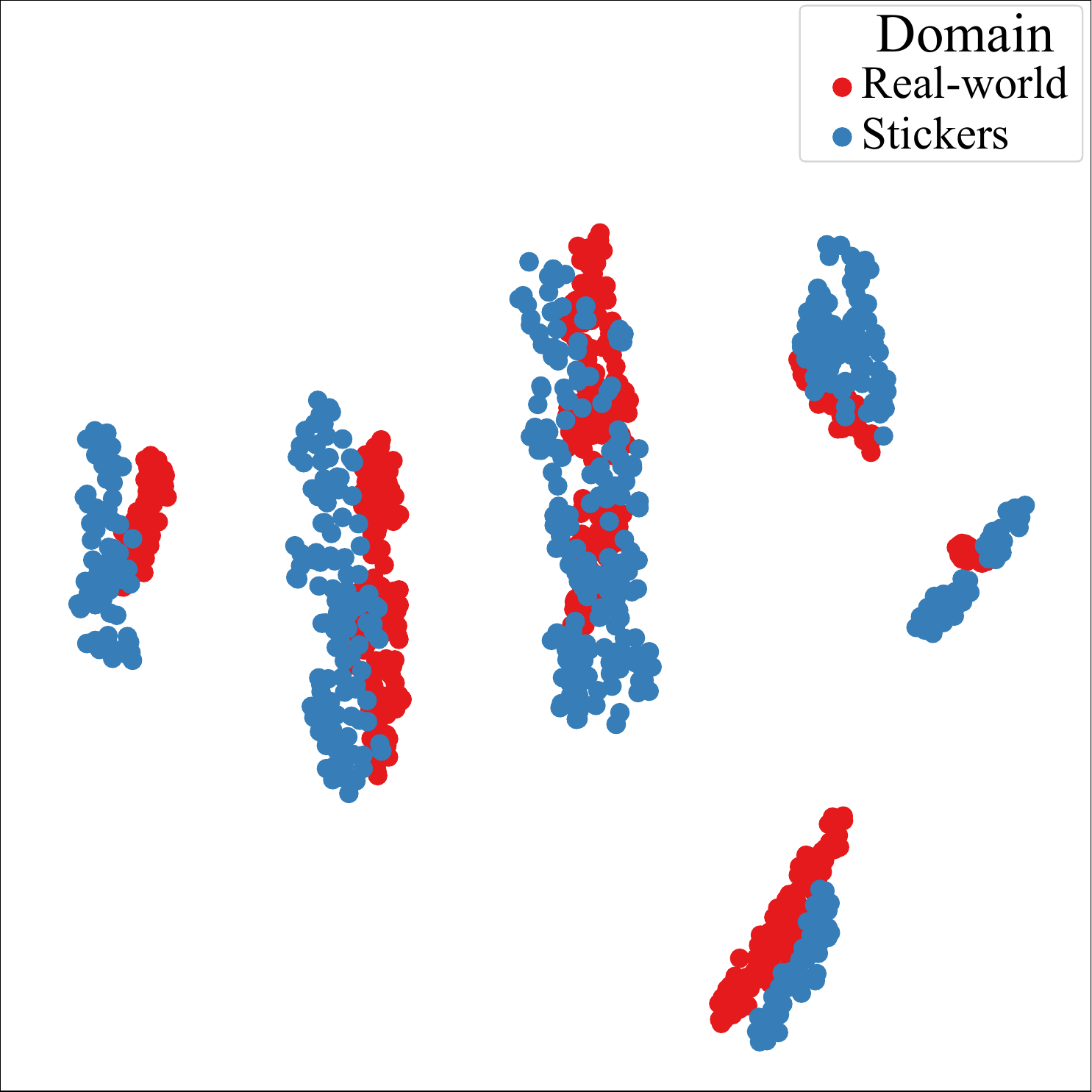} 
        \caption{Our method on E→S}
        \label{subpic:E2S_kadap}
    \end{subfigure}

    \caption{Visualization of cross-domain embedding versus baseline models using t-SNE \cite{tsne} on E→S task. Red and blue points represent the embedded representation of the source domain sample and the target domain sample, respectively.}
\end{figure}
Figure \ref{subpic:E2S_tpca} and Figure \ref{subpic:E2S_kadap} present a comparative analysis of the embedding space  generated by our proposed method and the baseline model, TGCA-PVT, for the domain adaptation (E→S).  
In Figure \ref{subpic:E2S_tpca} for TGCA-PVT, the source domain (red points) shows clear category-wise distinction, whereas the target domain (blue points) exhibits poor separation, indicating suboptimal classification performance in the target domain. 
In contrast, Figure \ref{subpic:E2S_kadap} illustrates the embedding space representation produced by our proposed method. Here, the red and blue points not only exhibit clearer class-wise separation but also demonstrate a more even distribution, indicating improved alignment between the source and target domains. 

\section{Conclusions}
In this paper, we proposed a new challenging task, Unsupervised Cross-Domain Visual Emotion Recognition (UCDVER), aiming to generalize visual emotion recognition from a source domain to a low-resource target domain in an unsupervised manner.  We identified two main challenges in UCDVER: significant variability in emotional expressions and a shift in affective distributions between domains. 
To address these challenges, we propose the Knowledge-aligned Counterfactual-enhancement Diffusion Perception (KCDP) framework consisting of two core components:   Knowledge-Alignment Diffusion 
Affective Perception (KADAP) and Counterfactual-Enhanced Language-Image Emotional Alignment (CLIEA).  We establish a new benchmark for UCDVER and achieve SOTA performance on many cross-domain scenarios. In the future, we will pay more attention to the universal emotion-transferring tasks.

\section*{Acknowledgments}
This research was partially supported by the National Natural Science Foundation of China (NSFC) (62306064 and U19A2059) and the Sichuan Science and Technology Program (2022ZHCG0008, 2023ZYD0165 and 2024ZDZX0011).   We appreciate all the authors for their fruitful discussions.
In addition, thanks are extended to anonymous reviewers for their insightful comments and suggestions.

{
    \small
    \bibliographystyle{ieeenat_fullname}
    \bibliography{main}
}

\end{document}